\documentclass[review]{elsarticle}

\usepackage{lineno,hyperref}
\modulolinenumbers[5]

\usepackage{amsmath,amssymb,amsthm}
\usepackage{breqn}
\usepackage{algorithm}
\usepackage{algorithmic}
\usepackage{flushend}

\usepackage{xcolor}

\usepackage{array}

\newtheorem{definition}{Definition}

\newcommand{\Probab}[1]{{P}({#1})}
\newcommand{\Pcond}[2]{\Probab{{#1}\mid{#2}}}

\renewcommand{\~}[1]{\overline{#1}}

\newcommand{\argmax}{\operatornamewithlimits{argmax}}

\newcommand*{\review}{\textcolor{black}}

\journal{Journal of Computational Science}

\begin{document}

\begin{frontmatter}


\title{Causal Data Science for Financial Stress Testing\footnote{
This is an extended version of the conference paper~\cite{GaoMR17}.
}}

\author[aff1]{Gelin Gao}
\author[aff1]{Bud Mishra}
\author[aff2]{Daniele Ramazzotti}

\address[aff1]{Courant Inst., New York University, New York, USA}
\address[aff2]{Stanford University, Stanford, California, USA}

\address{}

\begin{abstract}
The most recent financial upheavals have cast doubt on the adequacy of some of the conventional quantitative risk management strategies, such as VaR (Value at Risk), in many common situations. Consequently, there has been an increasing need for verisimilar financial stress testings, namely simulating and analyzing financial portfolios in extreme, \emph{albeit} rare scenarios. Unlike conventional risk management which exploits statistical correlations among financial instruments, here we focus our analysis on the notion of \emph{probabilistic causation}, which is embodied by Suppes-Bayes Causal Networks (SBCNs); SBCNs are probabilistic graphical models that have many attractive features in terms of more accurate causal analysis for generating financial stress scenarios. 

In this paper, we present a novel approach for conducting stress testing of financial portfolios based on SBCNs in combination with classical machine learning classification tools. The resulting method is shown to be capable of correctly discovering the causal relationships among financial factors that affect the portfolios and thus, simulating stress testing scenarios with a higher accuracy and lower computational complexity than conventional Monte Carlo Simulations. 
\end{abstract}


\end{frontmatter}


\section{Introduction}
\review{
This paper combines state-of-the-art algorithmic techniques from data science, machine learning, causality analysis and financial engineering to expose risks in financial markets via explicit adversarial scenarios constructed from historical data. It is more precise than traditional risk analysis via VaR and computationally more efficient than Monte Carlo simulation-based approaches. Thus, whereas it has been studied how to identify ``factors'' that negatively affect a financial asset (or portfolio) and how they are correlated, algorithms to generate a causally plausible adverse temporal trajectories of events have remained an unsurmountable challenge. Monte Carlo simulations using Bayes Network have found wide-spread use, but they do not produce an explainable framework, nor do they have attractive computational complexity. Human-expert generated scenarios can be augmented with rational explanation, but they lack consistency and scalability. Notwithstanding these challenges,
}
risk management has become a central part of world finance in the past decades
as financial regulators demand more quantitative risk assessments of financial entities. For example, the Basel Committee under International Bank of Settlements recommends that all banks maintain a minimum capital reserve. 

The proposed quantitative assessments are designed and implemented to mitigate the risk of insolvency: namely, the depletion of capital of financial entity to the point that it has to stop its operations. In accounting terms, for any financial entity, its account consists of assets, liabilities and net equity, where the famous accounting identity holds: \emph{Asset} $=$\emph{Liabilities} $+$ \emph{Net Equity}. The task of quantitative risk management is to calculate the amount of equity that has to be reserved so that the net equity will not drop to negative when potential risks materialize into actual losses \cite{duffie2005risk}. In other words, this excess capital reserve serves as a `risk buffer' that will absorb potential losses and prevent the financial entity from bankruptcy. Before the catastrophic financial crisis in 2008, the risk assessments were in general statistical risk measures like Value-at-Risk \cite{manganelli2001var}. The central idea behind statistical risk measure is: we assess the statistical distribution of our portfolio or balance sheet, and estimate the statistically large adverse moves. If our capital reserve is enough to cover such losses, then we can safely assume that we are statistically free from insolvancy. Value-at-Risk is the most widely used measure, which assesses usually the worst 1\% loss. Depending on different financial entities, hedge funds, banks or clearing houses, and on different financial instruments, stocks, bonds, or derivatives in their balance sheets, the specific methods of calculating such risk measures like VaR may vary, but generally they can be reasonably estimated by methods like Monte Carlo Simulation \cite{raychaudhuri2008montecarlo}. 

However, such conventional approaches became discredited when the recent events led to major financial catastrophes. For example, in the recent 2008 financial crisis, the reserves calculated the risk by using methods such as VaR which proved to be painfully inadequate. In a recent analysis of VaR before and during the financial crisis conducted by the Federal Reserve Board, the average bank profit-and-loss(PnL) did not exceed the bank VaR from December 2003 to April 2007, while the bank average PnL exceeded VaR six times from June 2007 to March 2008 \cite{FedVaRBackTest}. In other words, banks which maintained capital reserves equal to their Value-at-Risk would face on average six near bankruptcies during the crisis. What most intrigued the economists and political and social scientists, was the sheer lack of a single plausible causal explanation of these events -- which is thought to have ranged from (i) ``One eyed Scottish idiot!” (Jeremy Clarkson); (ii) ``Complex financial products; undisclosed conflicts of interest; the failure of regulators, the credit rating agencies, and the market itself.” \cite{Levin-CoburnReport}, Interest Rate Spreads, Emerging Markets: e.g., BRICS” -- and so on. Both the unusual abruptness and intuitive implausibility earned such scenarios the name, ``\emph{Black Swan Events\/}'' -- and many more. It also raised the question whether there is a theory of ``causality'' that can rigorously explain such events empirically from data -- we suggest that the machinery of model checking for a suitably expressive logic (e.g., PCTL Probabilistic Computational Tree Logic, a branching time propositional modal logic) provides just the right capabilities to succinctly specify and efficiently verify statements about such scenario. It derives its power from the way it combines logic, probability and reasoning about time.

More informally, these approaches could address the deeply-felt need for better regulation (and intervention) in the form of \emph{stress testing}. Stress testing refers to the analysis or simulation of the response of financial instruments or institutions, given intensely stressed scenarios that may lead to a financial crisis \cite{claessensi2013crisis}. For example, narrowly speaking, stress testing may model the response of a portfolio when Dow Jones suddenly drops by 5\%. The difference between stress testing and conventional risk management is that stress testing deliberately introduces an adversarial, albeit plausible event, which may be highly improbable but not implausible -- e.g., afore-mentioned \emph{black swan event\/} triggering an unforeseen scenario. Thus, stress testing must be capable of observing the response of financial instruments or institutions under extremely rare scenarios. Such scenarios must be deemed to be unlikely to be observed in conventional risk management, where the simpler system may fail to estimate a $99^{\rm th}$ percentile of the loss distribution, and subsequently leading to a claim that, with 99\% confidence level, a specific portfolio will perform well, giving a false sense of security. 


\subsection{\review{Our Contribution}}

Our core contribution is a stress testing method built on Suppes' causality structure and a novel algorithm to create and traverse Suppes Bayes Causal Networks (SBCN). Note that we had originally developed and applied this causality framework to study cancer progression, but had not explored its combination with ML (machine Learning), as here, for generating rare adversarial scenarios for stress testing. \review{ The integration of causality analysis with machine learning results in a novel and practical (albeit approximate) approach to risk analysis that is currently lacking in data science.} 
 
This paper evolves from Rebonato's use of Bayesian Network~\cite{rebonato2010coherent} as the core modeling technique, and extends beyond his method to combine the three stress testing scenario generation methods. Our scenario generation method samples from a conditioned Bayesian Network learned from historical data that is able to capture the causality structure between risk factors and financial assets. The advantage of this method is that the clear causal structure makes interpretation of stress testing results more intuitive. The method also incorporates machine learning tools to identify scenarios that are most detrimental to specific portfolios and reduce the computational complexity of sampling.  

\review{Our second contribution is in augmenting traditional factor models with causality analysis -- a challenging area in data science, especially, in finance and econometrics. For instance,}
after the 2008 Sub-prime Mortgage Crisis, many attempts have been taken to explore the causes of the gigantic crisis. Many attributed the crisis to very `direct' causes -- as suggested earlier, they would include: low-quality mortgage loans whose risk is concealed by securitization; derivatives like credit-default-swaps which helped support lending. Others presented more `indirect' causes: banks' capital requirements by Basel Accord, which encouraged securitization; long-term record low interest rates encouraged reckless borrowing. \cite{CrisisCause2008Dennis} People's interests in causality have grown tremendously since the crisis, since it is easier to understand cause and effects, than association or correlation, just like in natural sciences. However, causality structure is more than just cause and effects. In the explanations of the crisis mentioned above, we can already see the different, `direct' or `indirect' causes convolute together: capital requirements encouraged securitization and securitization hid risk; long-term low interest rate encouraged reckless borrowing, which led to the existence of low-quality mortgages. The past attempts admitted the `convoluted interactions' between causes, but failed to explore the actual complex causality structure. Nevertheless, the true discovery of the causality structure is crucial not only to the understanding of the interactions between causes and effects, but also to the generation of sound hypothetical scenarios. \review{The algorithmic framework presented here takes us one step closer to understanding various latent causal structures at play in a complex financial market.}

Our final contribution is in providing a practical and scalable implementation of financial causality analysis building on a theoretical foundation of causality with a rich and deep philosophical history. The start of modern causality theory is Scottish philosopher David Hume's regularity theory. The core of his theory is temporal priority, which means that causes always come before their effects, or in other words, causality follow a pattern of succession in time \cite{hume1793inquiry}. Following Hume, Judea Pearl's notion of intervention has laid the foundation of many modern computer algorithms for causal network inference. Intervention by Pearl implies that if we manipulate $c$ and nothing happens, then $c$ cannot be cause of $e$, but if a manipulation of $c$ leads to a change in $e$, then we may conclude that $c$ is a cause of $e$, although there might be other causes as well \cite{pearl2003causality}. Unrelatedly, Patrick Suppes proposed his notion of \emph{prima facie} cause which extends the ideas of Hume and Pearl, and this paper efficiently automates discovery of Suppes' prime facie causation to construct the causal Bayesian network of financial factors and assets. 

\review{To iterate an earlier point, the work presented here builds on our earlier work on cancer progression, but also addresses many practical challenges -- unique to financial data -- where computational efficiency is paramount,  but nontrivial.}

\subsection{\review{Road-map:}} This paper is organized as follow. Next section describes the background and related work. The section, following immediately, addresses theoretical foundations of our method and, in particular, it shows how combining the expressivity of Suppes-Bayes Causal Networks together with classical classification approaches can effectively capture the dynamics of financial stress testing. Section $4$ provides results
describing the accuracy (specificity and sensitivity) of our algorithm for the efficient inference and traversal of SBCNs from financial data and discusses its performance in-depth; it shows on realistic simulated data how our approach is preferable in comparison to the standard Bayesian methods. Section $5$ concludes the paper. 

\section{Literature Review}

\review{The emerging area of financial stress testing is still an embryonic field and has a relatively meager literature -- traditional data science approaches are not directly applicable; automation of the manual methods relying on domain expertise is mostly unformalized.
}

\subsection{\review{Stress Testing Literature in Finance}} Before the financial crisis, stress testing only enjoyed interest among advanced financial practitioners like risk managers and central bankers. Nevertheless, the severity of the global financial crisis and its unexpected nature suggested that a more extensive and rigorous use of stress testing methodologies would be crucial to reduce the occurrence of similar catastrophes. \cite{StressTestMario} Stress testing first emerged as banks' internal self-assessment of their financial soundness in the early 1990s. \cite{StressTestHistoryBOE} These stress tests were small-scale tests for individual banks to assess their own trading activities and balance sheets. Later, in 1996, the Basel I Market Risk Amendment required banks to develop stress tests as part of their internal models for the calculation of capital requirements for market risk. \cite{StressTestMario} In 2004, Basel II introduced requirement for credit risk stress testing by banks. Most recently, in 2011, the Federal Reserve began Comprehensive Capital Analysis and Review (CCAR) program which incorporated an annual bank stress test. \cite{StressTestHistoryBOE} The start of CCAR marks a nation-wide implementation of stress testing as a regular financial stability assessment. Stress testing thereby became one of the much debated topics in financial regulation and risk management. 

\subsection{\review{Stress Testing Literature in Data Science}} Recently, many different approaches have been developed to implement stress testing. In general, a stress testing procedure consists of two steps: $(i)$ \emph{generation of stress scenarios}, and $(ii)$ \emph{stress projections}. The first step generates the adversarial, albeit plausible stress scenario. The second step projects financial portfolios or banks' balance sheets onto the stress scenario and estimates the potential loss. In terms of stress scenario generation, the most direct method is the historical one, in which observed events from the past are used to test contemporary portfolios \cite{stress_testing_methods}. Some example historical scenarios used by practioners are: Black-Monday in 1987, Asian Crisis in 1997, and Financial Crisis in 2008. \cite{StressTestHistoryBOE} As an alternative, the event-based method has been proposed in order to quantify a specific hypothetical stress scenario subjectively, by domain experts, and then estimate the possible consequence of such event using macroeconomic and financial models \cite{stress_testing_methods}. To ensure a scenario is damaging to the portfolio, a portfolio-based method has been also studied in order to link scenarios directly with the portfolio \cite{stress_testing_methods}. To this extent, portfolio-based methods rely on Monte Carlo Simulation to identify the movements of risk factors that stress the given portfolio most severely.

However, all of these scenario generation methods have their own limitations. The historical approach is objective since it is based on actual events, but it is not necessarily relevant under the present conditions. The event-based hypothetical method is more relevant, but it relies intensively on expert judgment on whether a hypothetical event will be severely-damaging, albeit still plausible to occur. Sometime such judgment becomes difficult when the relationship between the underlying risk factors and the portfolio is unknown. Hypothetical methods have been blamed for their high degree of uncertainty. Practitioners sometimes find it hard to interpret the result of stress testing on hypothetical events since the probability of occurrence of the event is uncertain \cite{rebonato2010coherent}, and the construction of the hypothetical events are subjective. The portfolio-based method relies heavily on Monte Carlo Simulation, but brute force Monte Carlo Simulation is computationally inefficient especially when dealing with many risk factors. Also, portfolio-based methods are difficult to implement for nation-wide inter-bank stress testing like CCAR. 

To solve this problem, Rebonato et al. proposed a sampling approach based on Bayesian networks in \cite{rebonato2010coherent}, which naturally relied on correlations, but not causation. \review{Our work, presented here addresses this shortcoming.}

\section{\review{Method}}

\review{The underlying stress testing method builds on several ideas from a diverse sets of fields: Finance, Machine Learning, Causal Data Science and Algorithmics; we discuss these building blocks successively.}

\subsection{\review{Finance Theory}}
\review{Traditionally markets are thought to be \emph{efficient} and follow CAPM (Capital Asset Pricing Models), which assumes that return of an asset may be defined as follows:
\begin{equation}
r = R_{f} \, + \, \beta_{1}(K_{m}-R_{f}) \,+\alpha,
\end{equation}
where $r$ is the return of the asset, $R_f$ the risk free return (usually measured in terms of government treasury returns) and $K_{m}$ the market factor (measured as value-weighted market portfolio, similar to stock indexes). Such a model is of little interest in terms of stress testing of a portfolio of assets as all assets are equally correlated to the market and are affected similarly by any scenario.}

For our purposes, it is more meaningful to assume that the stocks are affected differently by different econometric factors and are causally intertwined. For example, we may adopt a common stock factor model, the \emph{Fama French Five Factor Model} \cite{fama1996multifactor}, where the return of the asset is defined as follows: 

\begin{equation}
r = R_{f} \, + \, \beta_{1}(K_{m}-R_{f}) \, + \, \beta_{2}SMB \, \\ + \, \beta_{3}HML \, + \, \beta_{4}RMW \,+ \, \beta_{3}CMA+\alpha.
\end{equation}

In the equation, 
\begin{itemize}
\item $r$ is the return of the asset; 
\item $R_{f}$ is the risk free return, usually measured in terms of government treasury returns; 
\item $K_{m}$ stands for market factor, measured as value-weighted market portfolio, similar to stock indexes; 
\item $SMB$ (Small Minus Big) stands for company size factor, measured by return on a diversified portfolio of small stocks minus the return on a diversified portfolio of big stocks;
\item $HML$ (High Minus Low) stands for company book-to-market $(B/M)$ ratio factor, measured by difference between the returns on diversified portfolios of high and low $B/M$ stocks, where $B/M$ is the ratio between company's book value to market value; 
\item $RMW$ (Robust Minus Weak) stands for company operating profitability factor, measured by the returns on diversified portfolios of stocks with robust and weak profitability and 
\item $CMA$ (Conservative Minus Aggressive) stands for company investment factor, difference between the returns on diversified
portfolios of low and high investment stocks, called conservative and aggressive \cite{fama1996multifactor}. 
\end{itemize}

\review{Consequently, the factors may be assumed to evolve temporally following embedded causal relationships. Their effects on the portfolio of stocks may be inferred by linearly regressing historical returns $r$, onto the five factors. However, we will also need to infer from data the temporal and probability-raising relations among the pairs of factors, which would indicate potentially genuine causal relations that affect the dynamics of the financial market. These provide the key ingredients of the plausible adversarial trajectories. 
}

\subsection{\review{Machine Learning Theory}}

We start with Machine Learning using Bayesian Graphical Models \cite{koller2009probabilistic}, popularly known as Bayesian networks, as a framework to assess stress testing, as previously done in this context by \cite{rebonato2010coherent}. Bayesian networks have long been used in biological modeling such as -omics data analysis, cancer progression or genetics \cite{beerenwinkel2007conjunctive,loohuis2014inferring,ramazzotti2015capri}, but their application to financial data analysis has been rare. Roughly speaking, Bayesian networks attempt to exploit the conditional independence among random variables, whether the variables represent genes or financial instruments. In this paper we adopt a variation of the traditional Bayesian networks as done in 
\cite{ramazzotti2016modeling,ramazzotti2016learning},
There Mishra and his co-authors have shown how constraining the search space of valid solutions by means of a causal theory grounded in Suppes' notion of probabilistic causation \cite{suppes1970probabilistic} can be exploited in order to devise better statistical inference algorithms. Also, by accounting for Suppes' notion of probabilistic causation, we ensure not only conditional independence but also \emph{prima facie} causal relations among variables, leading us to a better definition of the actual factors leading to risk. Moreover, through a maximum likelihood optimization scheme which makes use of a regularization score, we also attempt to only retain edges in the Bayesian network (graphically depicted as a directed acyclic graph, DAG) that correspond to only genuine causation, while eliminating all the spurious causes
\cite{caravagna2015algorithmic}.


Yet, given the inferred network, we can sample from it to generate plausible scenarios, though not necessarily adversarial or rare. In the case of stress testing, it is crucial to also account for rare configurations, for this reason, we adopt auxiliary tools from machine learning to discover random configurations that are both unexpected and undesired. 

Here, we expand the concept sketched above, starting with a background discussion of our framework, by describing the adopted Bayesian models and causal theories and we then show how classification -- proviso an inferred causal model like SBCN is available -- can effectively guide stress testing simulations. 

\subsubsection{Traditional Bayesian networks}
Informally, Bayesian networks are defined as a directed acyclic graph (DAG) $G = (V, E)$, in which each node $\in V$ represents a random variable to which is associated a conditional probability table, and each arc $\in E$ models a binary dependency relationship. The nodes induce an overall joint distribution that can be written as a product of the conditional distributions associated with each variable \cite{koller2009probabilistic}. In this paper, without any loss of generality, we restrict our attention to Bernoulli random variables with support in $\{0,1\}$. Specifically, we will consider as inputs for our analyses a dataset $D$ of $m$ observations over $n$ Bernoulli variables; we refer to the next subsections for a detailed description of the meaning of such variables. More details about Bayesian networks may be found in \cite{koller2009probabilistic}. 

\begin{figure}[!ht]
\center
\includegraphics[width=0.60\textwidth]{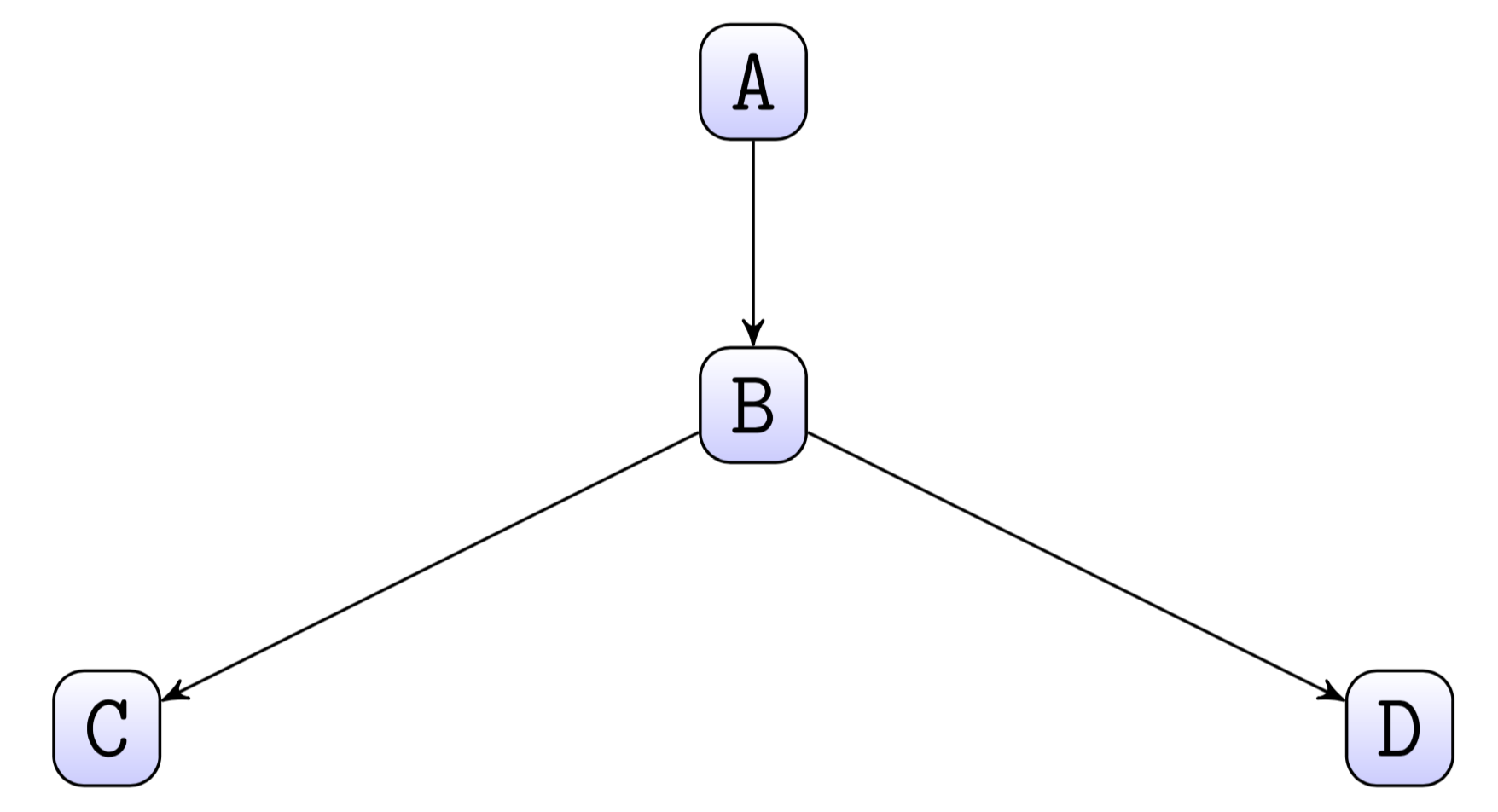}
\caption{Example of graphical structure of a Bayesian network with $4$ random variables.}
\label{fig:example_bayesian_network}
\end{figure}

Let us now consider as an example the Bayesian network shown in Figure \ref{fig:example_bayesian_network}, where $A$, $B$, $C$, and $D$ are $4$ random variables \review{(e.g., an econometric factor's value relative to a threshold)}
represented by four nodes, and the dependencies among the nodes are modeled by directed arcs. 
\review{Thus, a pertinent network could encode certain binary relations, such as correlations or causality, among Fama French Factors such as
$K_{m}$ (market factor, akin to stock indexes), $S$
(SMB, for company sizes), $H$ (HML,for company book-to-market $(B/M)$ ratio), $R$, (RMW, for company operating profitability), and $C$ (CMA, for company investment factor). The graph is defined by $V = \{K_m, S, H, R, C\}$ and $E \subseteq V \times V$.
}
Loosely speaking, the link $A \rightarrow B$ indicates that the knowledge of $A$ (the parent) influences the probability of $B$ (the child), or $A$ and $B$ are statistically dependent. Furthermore, for node $B$, node $A$ is called $B$'s parent and nodes $C$ and $D$ are called $B$'s children. More precisely, in the conditional probability tables related to the afore mentioned Bayesian network, the rows for node $B$ specifies how the knowledge of $A$ affects the probability of $B$ being observed. For example, let $A$ and $B$ be both binary random variables with support over $\{0,1\}$. Table \ref{table:example_conditional_probability_table} specifies the distribution of $B$ under the condition of $A$, and we can see clearly the effect of the parent on the child in this example. 

\begin{table}[h!]
\centering
\begin{tabular}{ | p{=1.0cm} | p{3.2cm} | p{3.2cm} | } 
\hline
 & \textbf{A = 0} & \textbf{A = 1} \\ 
\hline
\textbf{B = 0} & $P(B=0|A=0)=0.3 $ & $P(B=0|A=1)=0.4$\\ 
\hline
\textbf{B = 1} & $P(B=1|A=0)=0.7$ & $P(B=1|A=1)=0.6$\\ 
\hline 
\end{tabular}
\caption{Example of conditional probability table of node $B$ having node $A$ is unique parent.}
\label{table:example_conditional_probability_table}
\end{table}

One of the most significant feature of Bayesian network is the notion of \emph{conditional independence}. Simply speaking, for any node $X$ in a Bayesian network, given the knowledge of node $X$'s parents, $X$ is conditionally independent of all nodes that are not its children, or all its predecessors \cite{koller2009probabilistic}. For example, in the Bayesian network in Figure \ref{fig:example_bayesian_network}, node $C$ is conditionally independent of node $A$, when conditioned on node $B$ being fixed. The possibility of exploiting conditional dependencies when computing the induced distribution of the Bayesian network is a powerful property since it simplifies the conditional probability table tremendously. For example, the conditional probability table of node $C$, will not contain entries $P(C | A, B)$ since $P(C | A, B) = P(C | B)$, or node $C$ is independent of $A$ conditioned on $B$: $A \perp C | B$. 

In the context of stress testing, Rebonato \cite{rebonato2010coherent} suggests a subjective approach to constructing Bayesian networks. After carefully selecting a set of random variables as the nodes of the network, Rebonato proposes to subjectively connect the variables and assign the relevant conditional probability tables with the help of risk managers or other experts. Then with the inferred Bayesian network, reasoning about stressed events or simulation can be conducted. Please see \cite{rebonato2010coherent} for details. 

\subsubsection{\review{Suppes-Bayes Causal Networks \& Our Approach}}

Our framework builds upon many of Rebonato's intuitions but exploits our recent works on causality to address all the key problems, of which the subjective approach falls short. The subjective approach is handy under the condition of expert knowledge of the causal relationships of some variables. However, such reliance becomes unnatural when experts are confronted with random variables that are clearly beyond their expertise: for example, the relationship of unemployment and stock market performance, or more simply, the relationship of a pair of arbitrarily chosen stocks. Therefore, instead of completely abandoning the role of data in the construction of Bayesian network, here we adopt statistical inference algorithms that can learn both the structure and the conditional probability table of the Bayesian network from the data, which, in turn, can be further augmented by expert knowledge if deemed necessary. 

Thus, unlike \cite{rebonato2010coherent}, our stress testing approach builds on the foundation of Suppes-Bayes Causal Networks (SBCNs), which are not only more strictly regularized than the general Bayesian networks but also enjoys many other attractive features such as interpretability and refutability. SBCNs exploit the notion of probabilistic causation, originally proposed by Patrick Suppes \cite{suppes1970probabilistic}. 

In \cite{suppes1970probabilistic}, Suppes described the notion of \emph{prima facie causation}. A prima facie causal relation between any event $u$ and its effect $v$ is verified when the following two conditions hold: $(i)$ \emph{temporal priority} (TP), i.e., a cause happens before its effect and $(ii)$ \emph{probability raising} (PR), i.e., the presence of the cause raises the probability of observing its effect. 

\begin{definition}[Probabilistic causation,~\cite{suppes1970probabilistic}] \label{def:praising}
For any two events $u$ and $v$, occurring respectively at times $t_u$ and $t_v$, under the mild assumptions that $0 < \Probab{u}, \Probab{v} < 1$, the event $u$ is called a \emph{prima facie cause} of $v$ if it occurs \emph{before} and \emph{raises the probability} of $v$, i.e., 
\begin{equation}
\begin{cases}
(TP) \quad t_u < t_v \\
(PR) \quad \Pcond{v}{u} > \Pcond{v}{\~ u}\qquad
[\mbox{also} \equiv \Pcond{v}{u} > \Probab{v}.]
\end{cases}
\end{equation}
\review{where ${\~ u} \equiv \neg u$ is the Boolean complement of $u$ and corresponds to the event ``not $u$.''
Our reformulation\footnote{Note that:
$\Pcond{v}{u} > \Pcond{v}{\~ u} \equiv$
$\Pcond{v}{u}\Probab{u} + \Pcond{v}{u}(1 - \Probab{u}) > \Pcond{v}{u}\Probab{u} + \Pcond{v}{\~ u} (1 - \Probab{u})$
$\equiv \Pcond{v}{u} > \Probab{v}.$
} follows straightforward logic.
}
\end{definition}

The mathematical underpinnings of Probabilistic Causation are easily expressible in the logic below, which also allows efficient model checking in general. Thus enumerating complex \emph{prima facie} causes from data or probabilistic state transition models becomes feasible.
Thus, starting with a discrete time Markov chain (DTMC)\footnote{\review{A DTMC is a sequence of random variables following the Markov property, i.e., the probability distribution of future states only depends upon the present state \cite{markov1954algorithm}.}} -- a directed graph with a set of states, $S$, it is endowed (via labeling functions) with the atomic propositions true within them. It is possible to make the labeling probabilistic, so  that one may express that ``high market optimism'' may be false due to the fact that an adverse election results may be revealed with some small probability (e.g., depending on the status of a certain investigation).  The states are related pairwise by the transition probability. We also have an initial state from which we can begin a path (trajectory) through the system. Each state has at least one transition to itself or another state in $S$ with a non-zero probability. 

\review{A general framework for causality analysis is provided by model checking algorithms in PCTL (Probabilistic Computational Tree Logic) and has been explored in details by Mishra and his students~\cite{kleinberg2009temporal}. We start with a brief discussion on how Suppes' prima-facie causality can be formulated in PCTL, but then develop an efficient, albeit simplified, approach to financial stress testing using factor-models and SBCN (with pair-wise causality represented as edges in a graphical model) -- originally introduced by Mishra and his colleagues as a simplification. See \cite{kleinberg2009temporal,caravagna2015algorithmic,bonchi2015exposing}} More general, and computationally expensive (though tractable), approaches using PCTL will be explored in future research.

\begin{definition}[Probabilistic Computational Tree Logic, PCTL~\cite{ciesinski2004probabilistic}] \label{def:PCTL}
The types of formulas that can be expressed in PCTL are path
formulas  and  state  formulas.   State  formulas  express  properties  that must hold within a state, determined by how it is labeled with certain atomic propositions, while path formulas refer to sequences of states along which a formula must hold.
\begin{enumerate}
\item All atomic propositions are state formulas.
\item If $f$ and $g$ are state formulas, so are
\(\neg f \mbox{ and } f \wedge g. \)
\item If $f$ and $g$ are state formulas, and $t$ is a nonnegative integer or $\infty$, then
\(
f {\sf U}^{\leq t} g
\)
is a path formula.
\item If $f$ is  a  path  formula and $0 \leq p \leq 1$, then
\(
[f]_{>p}
\)
is a state formula.
\end{enumerate}
\end{definition}

\review{The syntax and the logic builds on standard propositional Boolean logic, but extends with various modes: the key operator is the metric ``until'' operator:
\(
f {\sf U}^{\leq t} g: 
\)
here, use of ``until'' means that one formula $f$ must hold at every state along the path until a state where the second formula $g$ becomes true, which must happen in less than or equal to
$t$ time units. Finally, we can add probabilities to these ``until''-like path formulas to make state formulas. Path quantifiers analogous to those in CTL may be defined by:
\({\sf A}f \equiv [f]_{\geq 1}\) [Inevitably $f$];
\({\sf E}f \equiv [f]_{> 0}\) [Possibly $f$];
\({\sf G}f \equiv f {\sf U}^{\leq \infty} {\sf false}\)[Globally $f$], and
\({\sf F} f \equiv {\sf true} {\sf U}^{\leq \infty} f
\) [Eventually $f$]. Formal semantics of the PCTL formul\ae\ may be found in~\cite{hansson1994logic}.}

\review{One can then say event $f$ ``probabilistic causes'' $g$, iff
\[
f \mapsto_{\geq p}^{\leq t} g \quad\Leftrightarrow\quad {\sf AG}\; (f \rightarrow {\sf F}_{\leq p}^{\leq t} g),
\]
for some suitable hyper-parameters $p$ probability and $t$ duration. Additional criteria (e.g., regularization) are then needed to separate spurious causality from the genuine ones -- as shown below. SBCN, thus, provides a vastly simplified, and yet practical, approach to causality, especially when explicit time is not recorded in the data.}

The notion of \emph{prima facie} causality was fruitfully exploited for the task of modeling cancer evolution in \cite{loohuis2014inferring,ramazzotti2015capri,caravagna2015algorithmic}, and the SBCNs were finally described for the first time in \cite{bonchi2015exposing} but, many of the basic ideas are already implicit in\cite{caravagna2015algorithmic}.


\begin{definition}[Suppes-Bayes Causal Network] \label{def:scn}
\emph{Let us consider an input cross-sectional dataset $D$ of $n$ Bernoulli variables and $m$ samples, the Suppes-Bayes Causal Network $SBCN = (V,E)$ subsumed by $D$ is a directed acyclic graph such that the following requirements hold:}
\item \emph{\textbf{[Suppes' constraints]} for each arc $(u \to v) \in E$ involving a prima facie relation between nodes $u,v \in V$, under the  mild assumptions that $0 < \Probab{u}, \Probab{v} < 1$}:
$$
\Probab{u} > \Probab{v} \quad \text{and} \quad \Pcond{v}{u} > \Pcond{v}{\neg u} \,.
$$
\item \emph{\textbf{[Sparsification]} let $E'$ be the set of  arcs satisfying the Suppes' constraints as before; among all the subsets of $E'$, the set of arcs $E$ is the one whose corresponding graph maximizes the likelihood of the data and of a certain regularization function $R(f)$:}
$$
E = \argmax_{E \subseteq E', G =(V,E)}  (LL(D | G) - R(f)) \,.
$$
\end{definition}

Intuitively, the advantage of SBCNs over general Bayesian networks is the following. First, with \emph{Temporal Priority\/}, SBCN accommodates the time flow among the nodes. There are obvious cases where some nodes occur before the other and it is generally natural to state that nodes that happen later cannot be causes (or parents) of nodes that happen earlier. Second, when learning general Bayesian networks, arcs $A \rightarrow B$ and $A \leftarrow B$ may sometimes be equally acceptable, resulting in an undirected arc $A-B$ (this situation is called \emph{Markov Equivalence\/} \cite{koller2009probabilistic}). For SBCNs, such a situation does not arise because of the temporal flow being irreversible. 
Third, because of the two constraints on the causal links, the SBCN graph is generally more sparse (has fewer edges) than the graph of general Bayesian networks with the final goal of disentangling spurious arcs, e.g., due to spurious correlations \cite{pearson1896mathematical}, from genuine causalities. 

\subsection{Machine Learning and Classification}
Even if SBCNs typically yield sparser DAGs than when we use Bayesian networks, the relations modeled  involve both positive and negative financial scenarios, but only in the latter financial stress may arise. Thus, the extreme events which are of key relevance for stress testing are still rare in the data and unlikely to be simulated in naively generated stress scenarios by sampling from the SBCN directly. Therefore, in this work we improve this basic model with several key ideas of classic machine learning, namely, feature classification. Recall that, in stress testing, we wish to target the unlikely, but risky scenarios. Specifically, when generating random sample from an SBCN to obtain possible scenarios, each node in the SBCN can take any value in its support according to its conditional probability table, generating different branches of scenarios. To narrow down the search space, we can classify each possible branch as leading to \emph{profitable} or \emph{lossy} scenarios, and if, the branch is classified as profitable, then random sampling is guided to very likely avoid that branch, thus focusing on events and causal relations that can be adversarial and risky, though uncommon. In this way, computation can be reduced significantly to discover the extreme events (see the next Sections for details). 
 
\subsection{\review{An efficient Implementation}}

The algorithm below, Algorithm~\ref{algo:sbcn}, encapsulates the earlier discussions.

Algorithm \ref{algo:sbcn} summarizes the inference approach adopted via SBCN. Given the above inputs, Suppes' constraints are verified (Lines $3$-$8$) to first construct a DAG. Then, the likelihood fit is performed by gradient ascent (Lines $9$-$20$), an iterative optimization technique that starts with an arbitrary solution to a problem (in our case an empty graph) and then attempts to find a better solution by incrementally visiting the neighborhood of the current one. If the new candidate solution is better than the previous one, it is considered in place of it. The procedure is repeated until the stopping criterion is matched.

In our implementation, the Boolean variable $=!${\it StoppingCriterion} is satisfied (Line $11$) in two situations: $(i)$ the procedure stops when we have performed a sufficiently large enough number of iterations or, $(ii)$ it stops when none of the solutions in $G_{neighbors}$ is better than the current $G_{fit}$, where $G_{neighbors}$ denotes all the solutions that are derivable from $G_{fit}$ by  removing or adding at most one edge. 

\newpage

\begin{algorithm}[ht]
\caption{Learning the SBCN \cite{bonchi2015exposing}}
\label{algo:sbcn}
\begin{algorithmic}[1]
\STATE{Inputs: $D$ an input dataset of $n$ Bernoulli variables and $m$ samples, and $r$ a partial order of the variables}
\STATE{Output: $SBCN(V,E)$ as in Definition $2$}
\STATE{\textbf{[Suppes' constraints]}}
\FORALL{pairs $(v,u)$ among the $n$ Bernoulli variables}
\IF{$r(v) \leq r(u)$ \AND $\Pcond{u}{v} > \Pcond{u}{\neg v}$}
\STATE{add the arc $(v,u)$ to $SBCN$.}
\ENDIF
\ENDFOR
\STATE{\textbf{[Likelihood fit by hill climbing]}}
\STATE{Consider $G(V,E)_{fit} = \emptyset$.}
\WHILE{$!StoppingCriterion()$}
\STATE{Let $G(V,E)_{neighbors}$ be the neighbor solutions of $G(V,E)_{fit}$.}
\STATE{Remove from $G(V,E)_{neighbors}$ any solution whose arcs are not included in $SBCN$.}
\STATE{Consider a random solution $G_{current}$ in $G(V,E)_{neighbors}$.}
\IF{$score_{REG}(D,G_{current})>score_{REG}(D,G_{fit})$}
\STATE{$G_{fit} = G_{current}$.}
\ENDIF
\ENDWHILE
\STATE{$SBCN = G_{fit}$.}
\RETURN $SBCN$.
\end{algorithmic}
\end{algorithm}

For more information about the algorithm, also refer to \cite{ramazzotti2015capri,bonchi2015exposing}. 

\subsection{\review{Our Contribution}}

\review{In this section we have shown how we integrated our earlier works on causality theory to produce an efficient implementation of a financial stress testing framework. However, since the implementation involves several hyper-parameters and different methods for regularization, the final embodiment requires additional empirical studies, which we describe earlier. For this purpose, tested and optimized it rigorously with a carefully selected synthetic financial model.
}


\section{\review{Results}}

\review{Next we describe our extensive comparative studies aimed at evaluating the statistical power of the frameworks that encompass the approaches of Rebonato (BNs) and ones proposed here (SBCNs) to perform stress testing. The other manual (expert-driven) approaches are out of the scope of this comparative studies for obvious reasons.}

Thus, the primary engines for stress testing are built with the generative models, which for our purposes are chosen to be one of of two kinds: Bayes Net (BN) or Causal Net (SBCN), but expected to behave differently based on the methods of model regularization: BIC (Bayesian Information Criterion) or AIC (Akaike Information Criterion) -- further constrained with or without bootstrapping.
Thus constructed, the resulting stress testing  algorithms may be investigated for performance, while paying specific attention to the problem of false discoveries (positive or negative). These results are succinctly visualized using ROC (Receiver Operating Characteristics). The data used for the analysis are simulated, as explained later.

\begin{figure}[!ht]
\center
\includegraphics[width=0.75\textwidth]{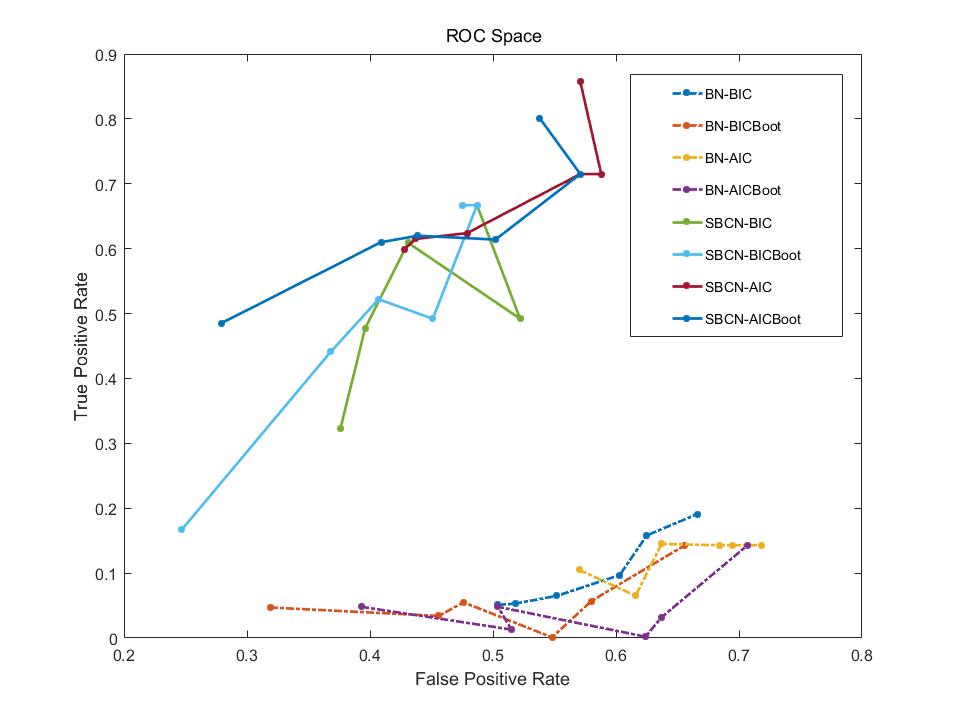}
\caption{Performance in the ROC Space, depicts the trade-offs between false positive rates and true positive rates. The better results lie in the upper-left corner of the graph where false positive rate is low and true positive rate is high.}
\label{fig:summary_results_learning}
\end{figure}

We summarize in Figure \ref{fig:summary_results_learning} the results of this analysis by interpolating and then smoothing out some kinks\footnote{Because of the data sparsity, the interpolation does not always lead to a smooth monotonic curves. } in order to obtain an ROC Space, whose $x$ axis represent the False Positive Rate and $y$ axis the True Positive Rate. ROC Space depicts the performance of the different methods we discussed on different sample sizes. By examining the plot, one can conclude that AICs generally have high true positive rates but also high false positive rates, as a result of its less stringent complexity penalty. In contrast, BICs generally have smaller false positive rates, but its true positive rates are also lower. Comparing the algorithms with and without bootstrapping, one can notice that the bootstrap procedure shifts the curves to the left. Still, the best performance lies in the data with the assumption of sparse relationships. Based on these results, we can conclude that with Bootstrapping and the assumption of sparse relationships, our algorithm is capable of recovering accurately the causal relationships in the data. 

\review{In order to provide further insights into these results, we describe in greater details: $(i)$ our simulation model that allows us to test the inferred results against the ground truth, $(ii.a)$ false-discovery analysis, $(ii.b)$ influence of information criteria (AIC and BIC) and $(ii.c)$ influence of bootstrapping. Finally, we describe the effect of Machine Learning in trajectory generation and projection from SBCN.}

\subsection{Training Data: Simulation and Evaluation with SBCN}
To assess the performance of the algorithm to infer the SBCNs and the quality of inferred Bayesian networks, a set of training data is developed with embedded causal relationships~\footnote{The vacuous case of ``no causality'' was not explored as it is not meaningful in the context of SBCN; this case was relegated to more general model checking approaches based on PCTL.}. If the algorithms, after `learning' a model from the training data, are capable of accurately recovering the causal relationships embedded in them, then comparable accuracy is to be expected on real data. 
To simulate the training data, we adopt the stock factor model, the \emph{Fama French Five Factor Model} \cite{fama1996multifactor}, introduced earlier.




To simulate the training data with embedded causal relationship, we linearly regress historical returns $r$, onto the five factors, and obtain the distribution of each factor coefficient and the empirical residual. We notice that a key characterization of an SBCN is an underlying temporal model of the causal relatas implicit in the network, namely the temporal priority between any pair of factors (represented by nodes of the SBCN) which are involved in a causal relationship. Therefore, the five factors described in our generative model are lagged with respect to the historical returns to comply with the temporal priority. Thus, 

\begin{equation}
r_{i,t} = \sum_{i,j} \beta_{i,j}f_{j,t-\mbox{lag}} + \epsilon,
\end{equation}
\review{where $f_{j, \cdot}$ is the $j$th factor's value at a time, properly ``lagged.''}

Then, the simple training data is simulated by randomly drawing the factor coefficients $\beta_{i,j}$ and residuals $\epsilon$ from the distribution we obtained from the linear regression, and apply these coefficients and residuals on a set of new factor data. Such historical data consists of daily series of five factors and returns of $10$ portfolios also constructed by Fama French, and of $10,000$ days. We use the first $5,000$ for regression and the other $5,000$ for simulation. 

In reality, many factors will present causal relationships among themselves. For example, some factors do not directly influence the asset, but affect the asset indirectly by its impact on other factors. Therefore, the simulated training data can be complicated by embedding spurious relationships also among factors. We linearly regress some factors on the other factors and simulate the training data in the same way. The choice of factors is arbitrary. In this paper, as an example, we regress the other four factors $SML$, $HML$, $RMW$ and $CMA$ on the market factor $K_{m}$. 

\begin{figure}[!ht]
\center
\includegraphics[width=0.60\textwidth]{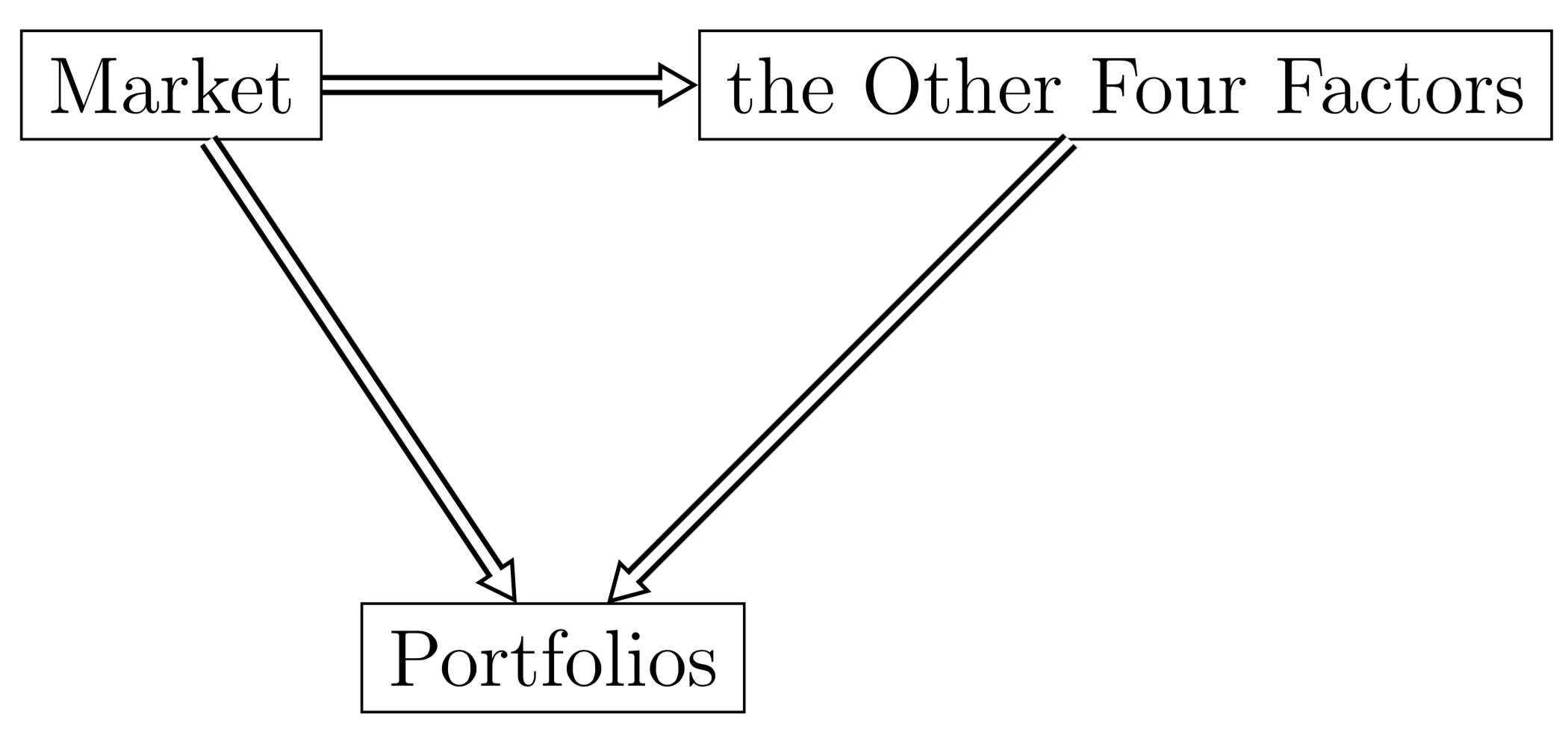}
\caption{Example of causal relationships described in the simulated training data.}
\label{fig:simulated_causal_relationships}
\end{figure}

Therefore, the causal relationships which are described in the simulated training data can be simplified as shown in Figure \ref{fig:simulated_causal_relationships}. 

Next we show results on $100$ independent random simulations generated on networks of $15$ nodes, i.e., $10$ stocks and $5$ factors with the generative model discussed in the previous Section. Each node represents a Bernoulli random variable taking binary values in $\{0,1\}$, where $1$ represents the stock or factor going up and $0$, the stock or factor going down. Specifically, the input of our learning task is a dataset $D \in \{0,1\}^{n\times m}$, an $n \times m$ binary matrix. Starting with such an input, we attempted to experiment with our learning algorithms previously described in \cite{ramazzotti2015capri} and \cite{bonchi2015exposing}. In particular, as in \cite{bonchi2015exposing}, we lacked explicitly observed time in the data, which are only cross-sectional. To overcome this problem we gave as a further input to our algorithm a topological ranking $r$ providing information about the temporal priority among the nodes. In interpreting these experiments, we set ranking as a proxy of time precedence among the factors influencing the stocks, i.e., in our model factors can cause stock moves but not the other way around. \review{This strategy results in removal of implausible spurious arcs going from stocks to factors, but without affecting any genuine constraints in the arcs among factors or among stocks.} 



\subsubsection{The Problem of False Discovery}
We first tested the performance of Algorithm \ref{algo:sbcn} on a training data of $10$ portfolios, $5$ factors and $5,000$ observations. On such settings, the algorithm was capable of recovering almost the whole set of embedded causal relationships with only $13$ false negatives, roughly, $33\%$ of total arcs; however, the number of false positives were unacceptably larger, reaching around $49\%$ of the total causal arcs obtained, thus requiring more attention to how the model was regularized. 

The explanation for this trend can be found in how the algorithm implements the regularization via Bayesian Information Criterion (BIC) \cite{schwarz1978estimating}, that is: 
\[
BIC = k\cdot\ln(N) - 2\ln(L),
\]
where $k$ is the number of arcs in the SBCN (i.e. number of causal relationships), $n$ is the number of observations of the data, and $L$ is the likelihood. The algorithm searches for the Bayesian network that minimizes the BIC. 

\begin{figure}[!ht]
\center
\includegraphics[width=0.80\textwidth]{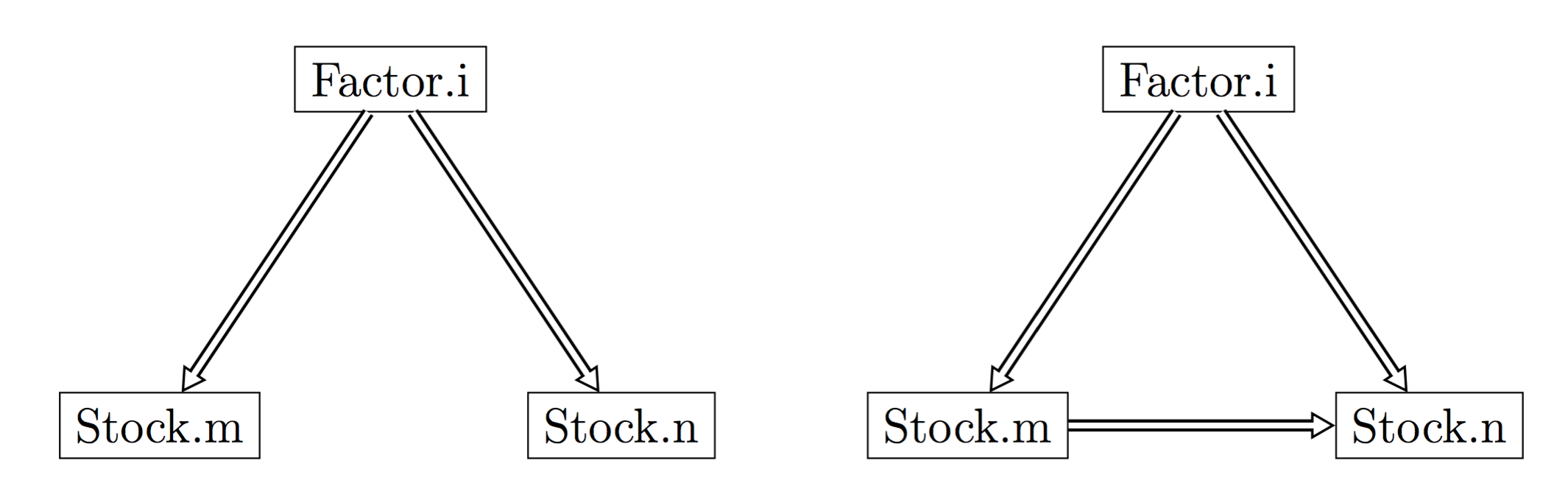}
\caption{The figure simplifies the true causal relationships (on the left), and the spurious relationships (on the right) emerging from the simulated data.}
\label{fig:simulated_spurious_relationships}
\end{figure}

For large number of observations, the maximum likelihood optimization ensures that asymptotically all the embedded relationships are explored and the most likely solution is recovered. However, maximum likelihood is known to be susceptible to over-fitting \cite{koller2009probabilistic}, especially when, as in our case, it deals with small sample size in the training data. Furthermore, in the training data, all the portfolios are assumed to depend on the same five factors, although with different coefficients, but very likely some portfolios will have very similar coefficients, resulting in co-movements across the portfolios. This co-movement will often induce correlations that affect the probability raising and thus the spurious prima facie causal relations, making these settings an interesting, and yet a very hard test case. See Figure \ref{fig:simulated_spurious_relationships}. 

\subsubsection{Sample Size and Information Criterion}
To reduce the spurious causalities, we recall some intrinsic properties of the information criteria. The Bayesian Information Criterion $BIC = k\cdot\ln(N) - 2\ln(L)$, not only maximizes the likelihood, but also penalizes the complexity of the model by the term $k\cdot\ln(N)$. For small sample sizes, BIC is generally biased towards simple models because of the penalty. However, for large sample size, BIC is willing to accept complex models. For additional discussion, see the details in \cite{koller2009probabilistic}. 

In our simulations we adopted a sample size of $5,000$ which is considerably large relative to the degree of freedom of the score function, thus inducing BIC to infer a relatively complex model with a number of unnecessary spurious arcs. Counter-intuitively, we could improve the solutions by using smaller sized data and letting the complexity penalty take a bigger effects in BIC score. This  strategy also addresses the non-stationarity in the data, an endemic problem for financial data. Following this intuition, we performed further experiments by reducing the original sample size of $5,000$ samples, which describes around $10$ years of data, in turn to $250$ and $500$, and we observed a significant reduction in the number of false positives, to $38\%$ and $40\%$ of total arcs respectively. However, at the same time, because of smaller sample size, the number of false negatives inevitably increased to more than $50\%$ of total arcs. 

To reconcile this dilemma, we next considered an alternative information criterion, the AIC, Akaike Information Criterion \cite{akaike1998information}, defined as in the following: 
\[
AIC = 2k - \ln(L).
\]

We notice that for AIC, the coefficient of $k$ is set to $2$, leading to definitely smaller factor than $\ln(N)$ of BIC when the sample size $N$ is large. For this reason, AIC supports the trend of accepting more complex models for given sample sizes than BIC. Applying AIC, the number of false negatives typically decreases, while the number of false positive gets larger. 

\subsubsection{Improving Model Selection by Bootstrapping}
So far we have described the different characteristics of two state-of-the-art likelihood scores while aiming to minimize the number of resulting false positive and false negative arcs in the inferred model. Specifically, we showed a trade-off where, because of their characteristics, the best results on large sample sizes is obtained using BIC, while for small sample sizes AIC is more effective, but neither of the two regularization schemes display a satisfactory trend. To improve their performance, we then examined a bootstrap \cite{efron1981nonparametric} procedure for model selection. 

The idea of bootstrap is the following: we first learn the structure and parameters of the SBCN as before, but we perform subsequently a re-sampling procedure where we sample with repetitions data from the dataset in order to generate a set of \emph{bootstrapped datasets}, e.g., $100$ times, and then we calculate the relative confidence level of each arc in the originally inferred SBCN, by performing the inference from each of the bootstrapped dataset and counting how many times a given arc is retrieved. In this way, we obtained a confidence level for any arc in the SBCN. 

We once again tested such an approach on our simulations and we observed empirically that the confidence level of spurious arcs are typically smaller than the confidence level for true causal relations. Therefore, a simple method of pruning the inferred SBCN to constrain for a given minimum confidence level is here applied. Such a threshold reflects the number of false positives that we are willing to include in the model, with higher thresholds ensuring sparser models. Here, we test our approach by requiring a minimum confidence level of $0.5$, i.e., any valid arc must be retrieved at least half of the times. 

We now conclude our analysis by showing in Tables \ref{table:results_exp_contincency_table_bn} and \ref{table:results_exp_contincency_table_suppes} the contingency tables resulting from our experiments both for Algorithm \ref{algo:sbcn} \textsc{(Table \ref{table:results_exp_contincency_table_suppes})} and for the standard likelihood fit method to infer Bayesian Networks (Table \ref{table:results_exp_contincency_table_bn}): 

\begin{table}[h!]
\centering
\begin{tabular}{ | p{1.1cm} | p{0.45cm} | p{0.45cm} | p{0.45cm} | p{0.45cm} | p{0.45cm} | p{0.45cm} | p{0.45cm} | p{0.45cm} | }
\hline
 & \multicolumn{2}{c|}{\textbf{BIC}} & \multicolumn{2}{c|}{\textbf{BICBoot}} & \multicolumn{2}{c|}{\textbf{AIC}} & \multicolumn{2}{c|}{\textbf{AICBoot}} \\
\hline
\textbf{Sample} & \textbf{FP} & \textbf{FN} & \textbf{FP} & \textbf{FN} & \textbf{FP} & \textbf{FN} & \textbf{FP} & \textbf{FN} \\
\hline
250 & 50.4 & 94.9 & 31.9 & 95.3 & 57.0 & 89.5 & 39.3 & 95.2 \\
\hline
500 & 51.8 & 94.7 & 45.6 & 96.6 & 61.6 & 93.5 & 51.5 & 98.7 \\
\hline
1000 & 55.2 & 93.5 & 47.6 & 94.5 & 63.7 & 85.5 & 50.4 & 95.2 \\
\hline
2500 & 60.3 & 90.3 & 54.8 & 99.9 & 68.4 & 85.7 & 62.4 & 99.8 \\
\hline
3500 & 62.5 & 84.2 & 58.0 & 94.3 & 69.5 & 85.7 & 63.7 & 96.8 \\
\hline
5000 & 66.6 & 80.9 & 65.6 & 85.7 & 71.8 & 85.7 & 70.7 & 85.7 \\
\hline
\end{tabular}
\caption{Contingency Table of the Performances by standard Bayesian Networks of Different Information Criteria and Sample Sizes.}
\label{table:results_exp_contincency_table_bn}
\end{table}

\begin{table}[h!]
\centering
\begin{tabular}{ | p{1.1cm} | p{0.45cm} | p{0.45cm} | p{0.45cm} | p{0.45cm} | p{0.45cm} | p{0.45cm} | p{0.45cm} | p{0.45cm} | }
\hline
 & \multicolumn{2}{c|}{\textbf{BIC}} & \multicolumn{2}{c|}{\textbf{BICBoot}} & \multicolumn{2}{c|}{\textbf{AIC}} & \multicolumn{2}{c|}{\textbf{AICBoot}} \\
\hline
\textbf{Sample} & \textbf{FP} & \textbf{FN} & \textbf{FP} & \textbf{FN} & \textbf{FP} & \textbf{FN} & \textbf{FP} & \textbf{FN} \\
\hline
250 & 37.6 & 67.7 & 24.7 & 83.3 & 42.8 & 40.1 & 27.9 & 51.5 \\
\hline
500 & 39.6 & 52.3 & 36.8 & 55.9 & 43.7 & 38.5 & 40.9 & 39.0 \\
\hline
1000 & 43.1 & 39.1 & 40.7 & 47.8 & 47.9 & 37.6 & 43.9 & 38.0 \\
\hline
2500 & 52.2 & 50.8 & 45.1 & 50.8 & 57.1 & 28.5 & 50.2 & 38.6 \\
\hline
3500 & 48.7 & 33.3 & 48.7 & 33.3 & 58.8 & 28.5 & 57.1 & 28.5 \\
\hline
5000 & 48.7 & 33.3 & 47.5 & 33.3 & 57.1 & 14.3 & 53.8 & 19.9 \\
\hline
\end{tabular}
\caption{Contingency Table of the Performances by Algorithm \ref{algo:sbcn} of Different Information Criteria and Sample Sizes.}
\label{table:results_exp_contincency_table_suppes}
\end{table}

Table \ref{table:results_exp_contincency_table_suppes} presents the results in terms of false positives (FP) and false negatives (FN) by Algorithm \ref{algo:sbcn} with the various methods on the training data with different information criteria, sample sizes, and whether Bootstrapping is applied. The trade-off between false positive rates and false negative rates usually is case-specific. We observe that, in general, the objective of such an approach is to correctly and precisely recover the true distribution underlying the training data. For this reason, unless differently specified for specific uses, there is not an overall preference toward either lower false positive or lower false negative. Therefore, we evaluate our methods by considering the sum of both false positive and false negative rates. This metric is biased toward a combination of relatively low FP and FN rather than the combination of very low FP and high FN and so on. By analyzing the results shown in Table \ref{table:results_exp_contincency_table_suppes}, we can clearly observe a trend where AIC with Bootstrapping on small sample datasets (i.e., $250$) and BIC with Bootstrapping on large sample datasets (i.e., $5,000$) produces the best results, which is in agreement with the discussion of the previous Section. Also, we observe that both for AIC without any bootstrapping on sample sizes of $250$ and BIC without any bootstrapping on sample size of $5,000$, the false positive rates are reduced without a significant increase in the false negative rates. 

We conclude by pointing out the significant increase in performance (both in terms of FP and FN) when using SBCNs in place of standard BNs (compare Tables \ref{table:results_exp_contincency_table_bn} and \ref{table:results_exp_contincency_table_suppes}). 

\subsubsection{Assumption of Sparse Relationships}
The resulting false positive rate may still seem relatively high. But, one important assumption is worth mentioning. In the training data, such high false positive rate derives from the fact that portfolios are dependent on the $5$ common factors, which very likely will induce co-movements. However, in the real data, such nested dependencies do not always occur, while a feature of sparse relationships appears frequently, and portfolios depend on distinctively small sets of factors. This assumption of sparsity can significantly improve the performance of the algorithm. 

To implement the assumption of sparsity, we deviate from the original Fama French five factor model. For simplicity, we generate data with sparse relationships using a random linear model with $10$ factor variables and $20$ stock variables. With $30\%$ probability, each stock variable is dependent linearly on one of the $10$ factor variables, so on average, each stock variable will be dependent on $3$ factor variables, which will likely be distinct from the dependent factor variables of other stock variables. Then we sample factor variable data from a normal distribution and compute the corresponding stock data using the linear model. 

Implementing this sparsity on a new set of purely random training data we obtain with Algorithm \ref{algo:sbcn} much better results, and, e.g., following the BIC with Bootstrapping method mentioned above, we obtain on small sample size ($250$ samples) $18.1\%$ false positive and $39.2\%$ false negative rates, while on large sample size ($2,500$ samples), we obtain $50.2\%$ false positive and $38.6\%$ false negative rates. 


\subsection{Practical Stress Testing}
In this Section we present how to assess stress testing scenarios given the inferred Suppes-Bayes Causal Network and we present the results on the simulated data. 

\subsubsection{Risk Management by Simulations}
After the inference of the SBCN, we perform Monte Carlo Simulation in the same way as conventional risk management, by drawing large number of samples to discover the worst $5\%$ scenarios as the value at risk (VaR). Nevertheless, here in stress testing, we are targeting the most extreme events, which have very low but nonzero probability of occurrence. Thus, they still can occur, for example, the $2008$ financial crisis or the most recent market reactions to \emph{BREXIT}. Therefore, when drawing samples from the network, we would like to reject the normal scenarios, and place more importance on the extreme events. To achieve this goal, when conducting random sampling, we classify each possible branch as profitable or risky, and if the branch is classified as profitable, then we will avoid that branch. 

\begin{figure}[!ht]
\center
\includegraphics[width=0.55\textwidth]{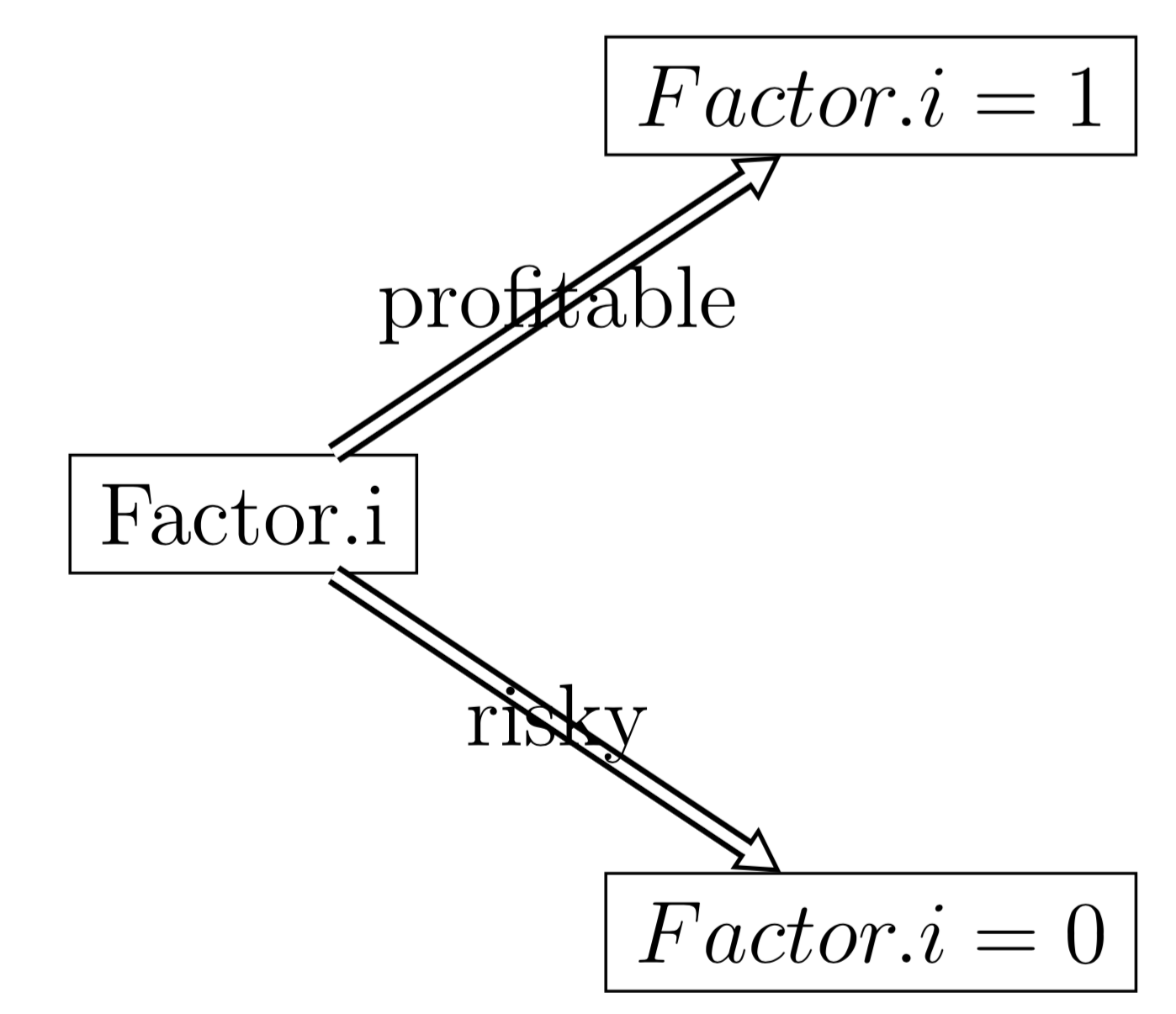}
\caption{Risk classification in our SBCN.}
\label{fig:risk_classification_sbcn}
\end{figure}

Figure \ref{fig:risk_classification_sbcn} represents a simple binary classification where for this factor only {\tt Factor.i} with value $0$ is considered risky and, hence, this scenario is the only one to be sampled. In this way, we target the extremely risky events and reduce computation. But, unlike conventional risk management, this approach does not allow us to estimate the probability of occurrence of the sampled extreme events, therefore we will not conclude a value at risk with certain confidence level. 

The simple binary classification with certain features is a standard machine learning problem. Here we explore a simple solution of such a task based on decision trees \cite{safavian1990survey}. A decision tree is a predictive models that maps features of an object to a target value of the object. Here, the features are the factors of interest, and the target value is whether the portfolio is prone to profit or loss. To perform classification, we first draw $1,000$ sample trajectories from the inferred SBCN. Then we construct a simple portfolio, which is long on all the stocks in the SBCN by the same amount, and calculate the Profit and Loss (P/L) of each observation. Here however, because the underlying SBCN depicts binary variables, exact Profit and Loss (P/L) statistics cannot be obtained. Instead, since the toy portfolio is long on all stocks by the same amount, the ratio of stocks that goes up is an approximate measure of risk. Of course, for continuous Bayesian network, Profit and Loss can be calculated directly. In the next step, we sort this measure, and denote the bottom-most $100$ scenarios as risky, and the rest as profitable. The $100$ `risky' scenarios contain at least $7$ stocks that fall. Then we consider $1,000$ samples each of them labeled as `risky' or `profitable.' In our experiments, we used the R `tree' package \cite{tree_r_package}. 

\begin{figure}[!ht]
\center
\includegraphics[width=0.75\textwidth]{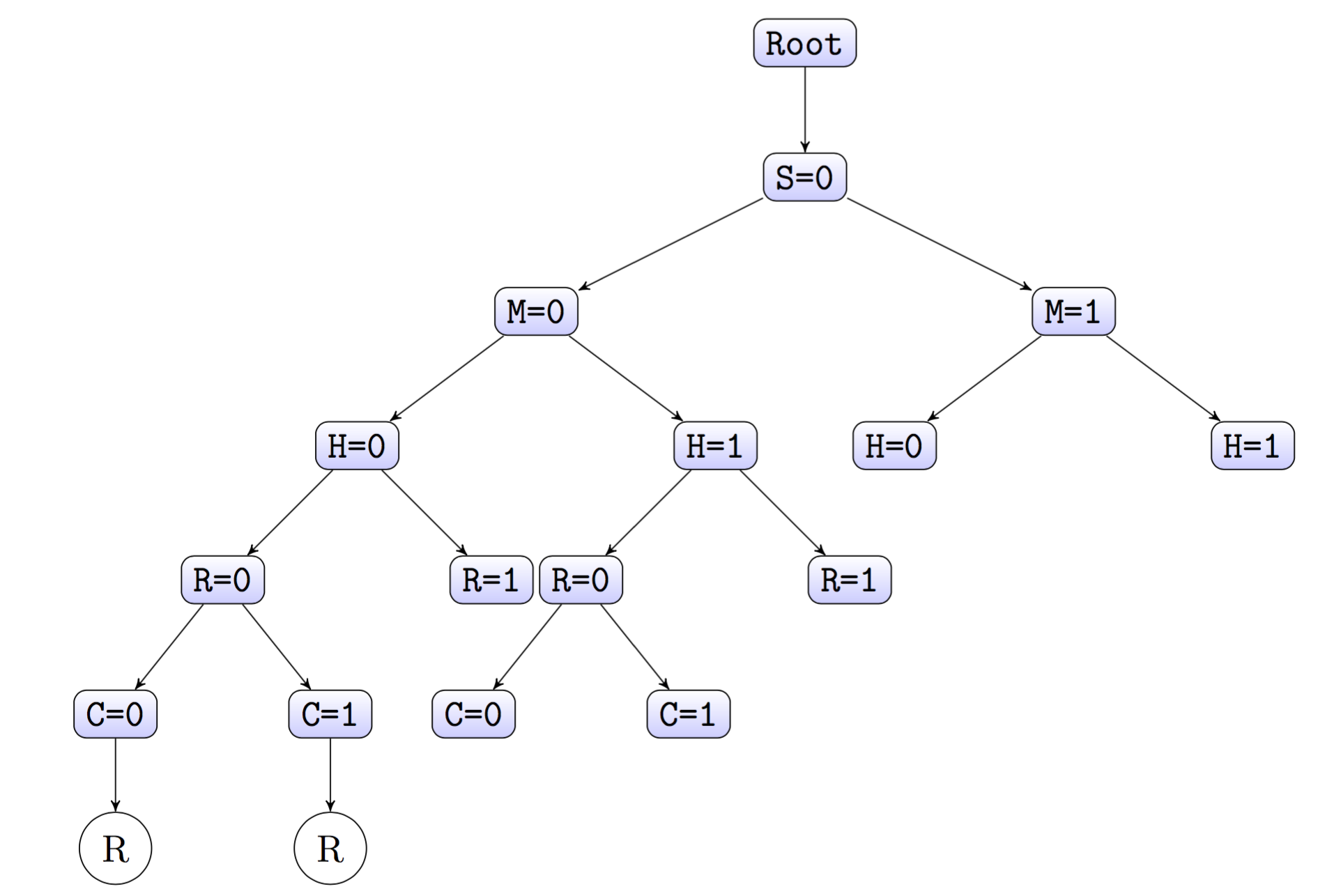}
\caption{Decision tree obtained from the SBCN.}
\label{fig:decision_tree_sbcn}
\end{figure}

Using the SBCN learned from the simulated training data, we obtain the following decision tree shown in Figure \ref{fig:decision_tree_sbcn}. 

In the decision tree of Figure \ref{fig:decision_tree_sbcn}, $S$ denotes factor $SMB$; $M$ denotes Market $K_{m}$; $H$ denotes $HML$; $R$ denotes $RMW$ and $C$ denotes $CMA$. Here we show only the left part of the entire computed decision tree, the subtree with $S=1$ is omitted, since the entire subtree with $S=1$ is classified as `Profitable,' which is not of interest for stress testing. In the tree, we identify two paths that are classified as `Risky.' Path $S=0$, $M=0$, $H=0$, $R=0$, $C=0$ and Path $S=0,M=0,H=0,R=0,C=1$. The paths classified are intuitive, since our example portfolio is long with equal amount invested over all $10$ stocks. Since $10$ stocks are generally positively dependent on the factors, most factors with $0$ values will likely induce a `Risky' path. For more complicated portfolios and real factors, such intuition cannot be easily found so we have to rely on the result of classification. 

\subsubsection{Scenario Generation and Results}
Given the tree of Figure \ref{fig:decision_tree_sbcn}, we then used the {\tt bnlearn} R package \cite{scutari2009learning} to sample from the SBCN. Given the network, we can simulate random scenarios, however, we wish not to simulate all of them, which will prove to be inefficient, but following the informations provided by the classification tree we choose the configurations which are likely to indicate risk to drive our sampling. For instance, we may pick the first path in the tree, which is $S=0$, $M=0$, $H=0$, $R=0$, $C=0$, and constrain the distribution induced by the SBCN. In order to avoid sampling the scenarios which are not in accordance with the path, we adjust the conditional probability table of the SBCN. Since we want paths with all five factors taking value 0, we set the conditional probability of these five factors taking value $1$ to $0$, and the conditional probability of factors taking value $0$ to $1$. In this way, the undesirable paths will be unlikely to be simulated, while the intrinsic distribution of how factors affect the stocks is still modeled. More sophisticated implementation based on this intuition are possible: e.g., using branch-and-bound, policy valuation, tree-search, etc, but we will leave this to future research. 

\begin{figure}[!ht]
\center
\includegraphics[width=0.60\textwidth]{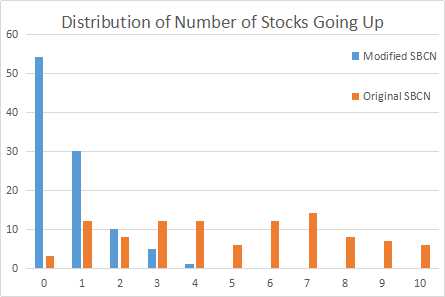}
\caption{Distribution of Number of Stocks Going Up.}
\label{fig:distributions_stocks_sbcn}
\end{figure}

Comparing the results of the simulations using the original SBCN and the one taking into account the decision tree, we show the distribution of the risk measure, the number of stocks that go up in the Figure \ref{fig:distributions_stocks_sbcn}. 

The number of stocks going up from $100$ samples generated by the original SBCN is roughly evenly distributed. At the same time, the $100$ samples generated by the modified SBCN contain no scenarios with more than $5$ stocks going up, and $84$ out of the $100$ samples have at most $1$ stock going up. We can clearly see that the modified SBCN places far more importance on the stressed scenarios, and in turn confirms the result of the classification algorithm by the decision tree. In this way, computational complexity involved in generating stressed scenarios can be improved tremendously. This kind of computational efficiency issues will be more critical when we move from a simple Bernoulli random variable to multi-categorical variables or continuous random variable. Therefore, with the same computing power, the modified SBCN makes it possible to generate more stressed scenarios, and observe how portfolios or other assets respond to stressed factors.







\section{Conclusion}
In summary, in this paper we develop a novel framework to perform stress testing combining Suppes-Bayes Causal Networks and machine learning classification. We learn SBCNs from data using Algorithm \ref{algo:sbcn} and assess the quality of the learned model by switching information criteria based upon sample sizes and bootstrapping. We then simulate stress scenarios using SBCNs, but reduce computation by classifying each branch of nodes in the network as `profitable' or `risky' using classification trees. For simplicity, the paper implements SBCNs with Bernoulli variables and simulates data using Fama French's Five Factor Model, but the logic of the problem is easily extended to deal with more practical situations. First of all, the SBCNs can accommodate more complicated variables (nodes). In addition to the factor based portfolios considered here, other factor models, or directly other financial and economic factors like foreign exchange rates, can also be included, and the accuracy of the model can ensure that the true causal relationships among the factors are discovered. In practice, variables like stock prices are continuous, thus, one can easily extend to these situations by adopting a hybrid SBCN, where the variables can take either discrete or continuous values, making it possible to represent precisely the values of the variables we are interested in. 

To use the model, the role of experts is still important. After learning the SBCN from data and applying classification, we can identify a number of stressed scenarios. However, we expect that some of these to be unacceptable for various unforeseen reasons, e.g., as those known to domain experts. These scenarios may be thought of as highly stressed with respect to the corresponding portfolio but they could prove to be less useful in practice. Therefore, experts can select from the identified stressed scenarios only the plausible ones, and discard the ones deemed to be flawed. Even in this case, we can perform simulations following the selected stressed trajectories in the SBCN and observe the reactions of the portfolios in these stressed scenarios of interest, and thus adjust the portfolios based on the reactions. Another direct usage of our approach is when experts have a particular candidate stress scenario in mind, which can be justified a priori; in this case one can skip the process of classification and directly adjust the SBCN {\em mutatis mutandis\/}. Therefore, simulations of the adjusted SBCN will also offer the reactions of the portfolio to this particular stressed scenario. 

We believe, based on our empirical analysis, that we have devised an efficient automated stress testing method using machine learning and causality analysis in order to solve a critical regulatory problem, as demonstrated by the algorithm's ability to recover the causal relationships in the data, as well as its efficiency, in terms of computation and data usage. \review{We plan to test our algorithms on real data to compare against human experts in a commercial setting, and based on our promising results with the simulated data, we are confident that the resulting platform will find a significant fraction (if not all) of the adversarial scenarios.}  

\section*{References}

\bibliography{references}

\begin{thebibliography}{10}
\expandafter\ifx\csname url\endcsname\relax
  \def\url#1{\texttt{#1}}\fi
\expandafter\ifx\csname urlprefix\endcsname\relax\def\urlprefix{URL }\fi
\expandafter\ifx\csname href\endcsname\relax
  \def\href#1#2{#2} \def\path#1{#1}\fi

\bibitem{GaoMR17}
G.~Gao, B.~Mishra, D.~Ramazzotti,
  \href{https://doi.org/10.1016/j.procs.2017.05.167}{Efficient simulation of
  financial stress testing scenarios with suppes-bayes causal networks}, in:
  International Conference on Computational Science, {ICCS} 2017, 12-14 June
  2017, Zurich, Switzerland, 2017, pp. 272--284.
\newblock \href {http://dx.doi.org/10.1016/j.procs.2017.05.167}
  {\path{doi:10.1016/j.procs.2017.05.167}}.
\newline\urlprefix\url{https://doi.org/10.1016/j.procs.2017.05.167}

\bibitem{duffie2005risk}
A.~J.~McNeil, R.~Frey, P.~Embrechts, Quantitative Risk Management: Concepts,
  Techniques and Tools, Princeton University Press, 2010.

\bibitem{manganelli2001var}
S.~Manganell, R.~F.Engle, Value at risk models in finance, European Central
  Bank Working Paper Series.

\bibitem{raychaudhuri2008montecarlo}
S.~Raychaudhuri, S.~J. Mason, R.~R. Hill, L.~Mönch, O.~Rose, T.~Jefferson,
  J.~W. Fowler, Introduction to monte carlo simulation, in: 2008 Winter
  Simulation Conference, 2008.

\bibitem{FedVaRBackTest}
J.~O’Brien, P.~J. Szerszen, An evaluation of bank var measures for market
  risk during and before the financial crisis, Finance and Economics Discussion
  Series.

\bibitem{Levin-CoburnReport}
C.~Levin, T.~Coburn, Wall Street and the Financial Crisis: Anatomy of a
  Financial Collapse, United States Senate Permanent Subcommittee on
  Investigations, 2011.

\bibitem{claessensi2013crisis}
S.~Claessens, M.~A. Kose, Financial crises: Explanations, types and
  implications, IMF Working Paper Series.

\bibitem{rebonato2010coherent}
R.~Rebonato, Coherent Stress Testing: a Bayesian approach to the analysis of
  financial stress, John Wiley \& Sons, 2010.

\bibitem{CrisisCause2008Dennis}
D.~McCuistion, D.~Grantham, Causes of the 2008 financial crisis.

\bibitem{hume1793inquiry}
D.~Hume, An inquiry concerning human understanding, Vol.~3, 1793.

\bibitem{pearl2003causality}
J.~Pearl, Causality: models, reasoning and inference, Econometric Theory
  19~(675-685) (2003) 46.

\bibitem{StressTestMario}
M.~Quagliariello, Stress-testing the Banking System : Methodologies and
  Applications, Cambridge University Press, 2009.

\bibitem{StressTestHistoryBOE}
K.~Dent, B.~Westwood, Stress testing of banks: An introduction, Bank of England
  Quarterly Bulletin 2016 Q3.

\bibitem{stress_testing_methods}
{Committee on the Global Financial System}, Stress testing at major financial
  institutions: survey results and practice (2005).

\bibitem{fama1996multifactor}
E.~F. Fama, K.~R. French, Multifactor explanations of asset pricing anomalies,
  The journal of finance 51~(1) (1996) 55--84.

\bibitem{koller2009probabilistic}
D.~Koller, N.~Friedman, Probabilistic graphical models: principles and
  techniques, MIT press, 2009.

\bibitem{beerenwinkel2007conjunctive}
N.~Beerenwinkel, N.~Eriksson, B.~Sturmfels, Conjunctive bayesian networks,
  Bernoulli (2007) 893--909.

\bibitem{loohuis2014inferring}
L.~O. Loohuis, G.~Caravagna, A.~Graudenzi, D.~Ramazzotti, G.~Mauri,
  M.~Antoniotti, B.~Mishra, Inferring tree causal models of cancer progression
  with probability raising, PloS one 9~(10) (2014) e108358.

\bibitem{ramazzotti2015capri}
D.~Ramazzotti, G.~Caravagna, L.~Olde~Loohuis, A.~Graudenzi, I.~Korsunsky,
  G.~Mauri, M.~Antoniotti, B.~Mishra, Capri: efficient inference of cancer
  progression models from cross-sectional data, Bioinformatics 31~(18) (2015)
  3016--3026.

\bibitem{ramazzotti2016modeling}
D.~Ramazzotti, A.~Graudenzi, G.~Caravagna, M.~Antoniotti, Modeling cumulative
  biological phenomena with suppes-bayes causal networks, arXiv preprint
  arXiv:1602.07857.

\bibitem{ramazzotti2016learning}
D.~Ramazzotti, M.~S. Nobile, M.~Antoniotti, A.~Graudenzi, Learning the
  probabilistic structure of cumulative phenomena with suppes-bayes causal
  networks, arXiv preprint arXiv:1703.03074.

\bibitem{suppes1970probabilistic}
P.~Suppes, A probabilistic theory of causality, North-Holland Publishing
  Company Amsterdam, 1970.

\bibitem{caravagna2015algorithmic}
G.~Caravagna, A.~Graudenzi, D.~Ramazzotti, R.~Sanz-Pamplona, L.~De~Sano,
  G.~Mauri, V.~Moreno, M.~Antoniotti, B.~Mishra, Algorithmic methods to infer
  the evolutionary trajectories in cancer progression, Proceedings of the
  National Academy of Sciences 113~(28) (2016) E4025--E4034.

\bibitem{markov1954algorithm}
A.~Markov, N.~Nagorny, Algorithm theory, Trudy Mat. Inst. Akad. Nauk SSSR 42
  (1954) 1--376.

\bibitem{kleinberg2009temporal}
S.~Kleinberg, B.~Mishra, The temporal logic of causal structures, in:
  Proceedings of the Twenty-Fifth Conference on Uncertainty in Artificial
  Intelligence, AUAI Press, 2009, pp. 303--312.

\bibitem{bonchi2015exposing}
F.~Bonchi, S.~Hajian, B.~Mishra, D.~Ramazzotti, Exposing the probabilistic
  causal structure of discrimination, International Journal of Data Science and
  Analytics 3~(1) (2017) 1--21.

\bibitem{ciesinski2004probabilistic}
F.~Ciesinski, M.~Gr{\"o}{\ss}er, On probabilistic computation tree logic, in:
  Validation of Stochastic Systems, Springer, 2004, pp. 147--188.

\bibitem{hansson1994logic}
H.~Hansson, B.~Jonsson, A logic for reasoning about time and reliability,
  Formal aspects of computing 6~(5) (1994) 512--535.

\bibitem{pearson1896mathematical}
K.~Pearson, Mathematical contributions to the theory of evolution.--on a form
  of spurious correlation which may arise when indices are used in the
  measurement of organs, Proceedings of the royal society of london
  60~(359-367) (1896) 489--498.

\bibitem{schwarz1978estimating}
G.~Schwarz, et~al., Estimating the dimension of a model, The annals of
  statistics 6~(2) (1978) 461--464.

\bibitem{akaike1998information}
H.~Akaike, Information theory and an extension of the maximum likelihood
  principle, in: Selected Papers of Hirotugu Akaike, Springer, 1998, pp.
  199--213.

\bibitem{efron1981nonparametric}
B.~Efron, Nonparametric estimates of standard error: the jackknife, the
  bootstrap and other methods, Biometrika 68~(3) (1981) 589--599.

\bibitem{safavian1990survey}
S.~R. Safavian, D.~Landgrebe, A survey of decision tree classifier methodology.

\bibitem{tree_r_package}
B.~Ripley, \href{https://CRAN.R-project.org/package=tree}{Tree: Classification
  and Regression Trees}, r package version 1.0-37 (2016).
\newline\urlprefix\url{https://CRAN.R-project.org/package=tree}

\bibitem{scutari2009learning}
M.~Scutari, Learning bayesian networks with the bnlearn r package, arXiv
  preprint arXiv:0908.3817.

\end{thebibliography}

\end{document}